\documentclass[]{article}
\usepackage{geometry}
\usepackage{footmisc}

\usepackage[backend=biber]{biblatex}
\usepackage{authblk}
\usepackage{graphicx}
\usepackage{epstopdf}
\usepackage{multirow}
\usepackage{booktabs}
\usepackage{amsmath}
\usepackage{amssymb}
\usepackage{amsfonts}
\usepackage{bbm}
\usepackage{lmodern}
\usepackage{titlesec}
\usepackage{hyperref}
\usepackage{dblfloatfix}
\usepackage{threeparttable}
\usepackage{subcaption}

\titleformat{\subsubsection}[runin]
  {\normalfont\normalsize\bfseries}{\thesubsubsection}{1em}{}
\titlespacing*{\subsubsection}{0pt}{0.7ex}{1.5ex plus .2ex}

\title{Estimating Carotid Pulse and Breathing Rate from Near-infrared Video of the Neck}
\author[1]{Weixuan Chen}
\author[1]{Javier Hernandez}
\author[1]{Rosalind W. Picard}

\affil[1]{The Media Lab, Massachusetts Institute of Technology, Cambridge, MA}

\date{May 2018}

\begin{document}


\maketitle

\begin{abstract}
\textit{Objective:} Non-contact physiological measurement is a growing research area that allows capturing vital signs such as heart rate (HR) and breathing rate~(BR) comfortably and unobtrusively with remote devices. 
However, most of the approaches work only in bright environments 
in which subtle photoplethysmographic and ballistocardiographic signals can be easily analyzed and/or require expensive and custom hardware to perform the measurements.

\textit{Approach:} This work introduces a low-cost method to measure subtle motions associated with the carotid pulse and breathing movement from the neck using near-infrared (NIR) video imaging. A skin reflection model of the neck was established to provide a theoretical foundation for the method. In particular, the method relies on template matching for neck detection, Principal Component Analysis for feature extraction, and Hidden Markov Models for data smoothing.

\textit{Main Results:} We compared the estimated HR and BR measures with ones provided by an FDA-cleared device in a 12-participant laboratory study: the estimates achieved a mean absolute error of 0.36 beats per minute and 0.24 breaths per minute under both bright and dark lighting.

\textit{Significance:} This work advances the possibilities of non-contact physiological measurement in real-life conditions in which environmental illumination is limited and in which the face of the person is not readily available or needs to be protected. Due to the increasing availability of NIR imaging devices, the described methods are readily scalable. 
\end{abstract}

%
%
%
%
\renewcommand*{\thefootnote}{\arabic{footnote}}

\section{Introduction}
Measuring physiological signals such as heart rate and breathing rate is critical for early diagnosis of chronic cardiovascular and respiratory diseases as well as their long-term monitoring. However, some of the most frequently used devices require attaching irritating electrodes on the skin and/or adding cumbersome straps around the torso. In addition, some of these devices have wires and require maintenance (e.g., replacing electrodes, charging batteries) which limit their use in daily life monitoring and often lead to high attrition rates. Moreover, these devices are not appropriate for certain settings in which the skin is too sensitive (e.g.,~neonates, burn patients) or in which a high level of unobtrusiveness is desired~(e.g.,~sleep monitoring). 

\begin{figure}[!ht]
	\centering
	\includegraphics[width=0.95\linewidth]{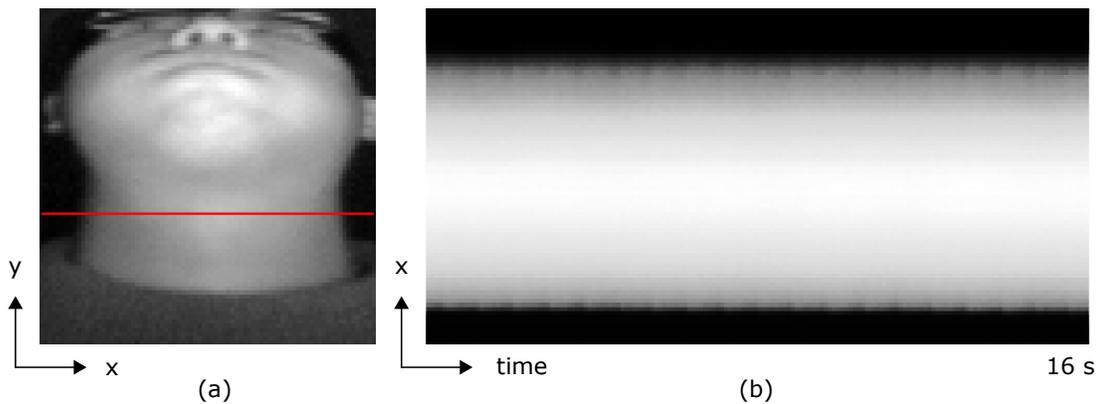}
	\caption{Carotid pulse visualization. (a) The first frame of a NIR video with a red scan line on the neck. (b) The scan line plotted over time, which shows a subtle crenation along both edges corresponding to the carotid pulse.}
	\label{pulseviz}
\end{figure}

To help minimize the previous challenges, researchers have explored measurement methods that can capture physiological signals remotely and without any skin contact.
Some of the most commonly used methods include but are not limited to thermal imaging of faces~\cite{Garbey2007,Fei2010,Cho2017}, wireless/radar signals bouncing on torsos~\cite{Adib2015}, and RGB videos of faces in bright conditions~\cite{sun2015ppg,Balakrishnan2013}. In contrast, this work uses low-cost NIR imaging devices to track subtle carotid pulse and breathing movements from the neck (see Fig. \ref{pulseviz}). Due to the increasing popularity of such devices, we believe this configuration complements previously explored methods and allows expanding the sensing opportunities to other real-life scenarios. In particular, we believe the proposed approach offers a good alternative in settings in which face-facing cameras may not be possible (e.g.,~monitoring drivers), the face of the person may want to be protected (e.g., ~workplace monitoring), and/or the illumination of the environment may be too dynamic or absent (e.g.,~sleep monitoring). 

The work is organized as follows. First, we describe relevant previous work in the context of non-contact physiological measurement. Second, we establish a skin reflection model of the neck and derive how the physiological signals can be separated from the other motion sources. Third, we describe the apparatus and experimental setup used in this work. Fourth, we describe the proposed methods to estimate HR and BR from NIR video. Fifth, we describe the results and discuss the main findings. Finally, we provide some concluding remarks and lines of future work.

\section{Previous Work}


Non-contact physiological measurement methods have significantly advanced during recent years. For instance, physiological parameters such as HR and BR have been accurately extracted from RGB facial videos in which subtle color changes of the skin caused by blood circulation can be amplified and analyzed \cite{verkruysse2008remote,Poh2010b,Poh2011b,DeHaan2013,Tarassenko2014,Wang2016b}. Similar metrics have also been extracted by analyzing subtle face motions associated with the blood ejection into the vessels (a.k.a., ballistocardiography) \cite{Balakrishnan2013} as well as more prominent chest volume changes during breathing \cite{Tan2010,Philips2011,Janssen2016}. Most of the previous studies leverage pervasive RGB cameras such as webcams that require a bright environment to work well \cite{Wu2012a}. However, this requirement may not be possible in many real-life scenarios such as sleep monitoring, dark office spaces, and night surveillance. 

To help address this problem, researchers have explored the use of other sensing modalities. For instance, microwave Doppler radar and thermal imaging have been used for non-contact HR and BR measurements~\cite{Greneker1997,Garbey2007,Fei2010,Cho2017}, and laser Doppler vibrometry has been used to remotely track the carotid pulse \cite{Rohrbaugh2006}. However, measuring these modalities usually requires expensive and custom hardware which is not readily available. 
A less expensive alternative involves using NIR video monitoring which has recently grown in popularity with the increasing interest in motion sensing and its potential uses in virtual and augmented reality (e.g., Intel RealSense, Leap Motion).
Using this modality\footnote{Here the modality refers specifically to raw NIR imaging. There are also methods estimating depth maps in NIR for physiological sensing \cite{Xia2012,Yang2015c}, but they rely on structured light sources or time-of-flight sensors associated with extra cost and complexity.}, researchers have analyzed videos of torsos to diagnose obstructive sleep apnea \cite{Ching2006}, and have observed pulsatile signals associated with arterial and breathing activity on the arm and the lower leg \cite{Such1997,Wieringa2005}. However, none of these previous methods directly estimated and evaluated physiological metrics from NIR frames. One exception is the works of van Gastel~et~al., which considered NIR face videos to estimate HR and BR \cite{VanGastel2015,VanGastel2016}. In their study, they observed that the relative photoplethysmographic amplitude was significantly reduced in NIR compared to visible light and, therefore, they needed to use three NIR wavelengths (hence three NIR cameras with different optical filters) to appropriately separate pulse-induced intensity variations from motion artifact. Similarly, our work also considers NIR video to estimate HR and BR. Instead of focusing on the face or the torso areas, we focus on the neck which may be more convenient in certain experimental conditions in which front facing cameras are not feasible. In addition, we focus on subtle cardiorespiratory motions instead of photoplethysmographic changes which allow us to rely on a single NIR wavelength. 


\begin{figure}[!bth]
	\centering\includegraphics[width=0.95\linewidth]{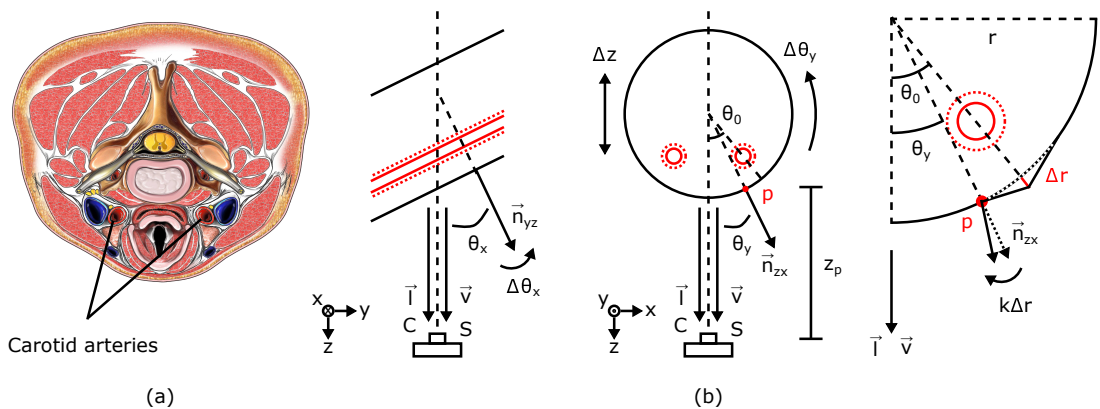}
	\caption{(a) Cross section of the human neck showing the location of the carotid arteries (b) Geometric relationships among the light source $S$, the camera $C$, and all the potential motion sources of the neck.}
	\label{phong}
\end{figure}

\section{Skin Reflection Model of the Neck}

Fig. \ref{phong} (a) shows a cross section of the neck and the location of the carotid arteries which can be simplified to a cylindrical model (Fig. \ref{phong} (b)). With each cardiac cycle, the heart pumps blood to the periphery, and causes palpable pulsatile changes in the carotid arterial diameter, reflected as subtle skin deformations $\Delta r(t)$ along the two sides of the neck. To model these deformations in video and its relationship with lighting, 
we adopted the Blinn-Phong reflection model. Compared with other models commonly used for modeling skin reflection such as the Lambert-Beer law and Shafer's dichromatic reflection model, the Blinn-Phong reflection model is better at illustrating geometric relationships.

Assuming a surface point $p$ on the neck illuminated by a single NIR light source $S$, its illumination captured by a NIR camera $C$ can be expressed as 
\begin{equation}
	I_p(t)=I_a k_a+I_i(t)(k_d(\vec{l}\cdot\vec{n}(t))+k_s(\vec{n}(t)\cdot\vec{h})^{\alpha})+I_n(t)
\end{equation}
It consists of four terms:

\begin{enumerate}
	\item $I_a k_a$ is the ambient light intensity, which is the result of multiple reflections from walls and objects. $I_a$ is the ambient component of the light source, and $k_a$ is the ambient reflection coefficient. The whole term is usually considered to be constant for a particular object and uniform at every point.
	\item $I_i(t)k_d(\vec{l}\cdot\vec{n}(t))$ is the diffuse reflection intensity, associated with the absorption and
	scattering of the light in skin-tissues. $I_i(t)$ is the intensity of the light source after distance attenuation. $k_d$ is the diffuse reflection coefficient, which depends on the nature of the material (skin) and the wavelength of the incident light. In theory, there is a pulsatile component in $k_d$ similar to photoplethysmography due to the variations of hemoglobin absorption. However, in NIR hemoglobin absorption of the skin is one order of magnitude smaller than dermal scattering and two orders of magnitude smaller than melanin absorption \cite{Mendenhall2015}, causing the component to be much weaker than in visible light \cite{VanGastel2015}. Thus we assume $k_d$ to be a constant. $\vec{l}$ is the direction vector from point $p$ toward the light source. $\vec{n}$ is the surface normal at $p$.
	\item $I_i(t)k_s(\vec{n}(t)\cdot\vec{h})^{\alpha}$ is the specular reflection intensity, which is a mirror-like light reflection from the skin surface. $k_s$ is the specular reflection coefficient, usually taken to be a material-dependent constant. $\vec{h}$ is the halfway vector defined as $\vec{h} = (\vec{l}+\vec{v})/2$, in which $\vec{v}$ is the direction vector from point $p$ toward the camera. As we assume $S$ is a distant light source at the same location as $C$, $\vec{h}=\vec{l}=\vec{v}$. $\alpha$ is the shininess coefficient controlling the strength of specular highlight.
	\item $I_n(t)$ is the quantization noise of the camera sensor.
\end{enumerate}

The movement of the neck involves rigid motions and non-rigid motions. Assuming a Cartesian coordinate system with its $z$-axis parallel to $\vec{l}$ and its $y$-axis perpendicular to the neck cross section, the rigid motions will include translations along three axes $\Delta x(t)$, $\Delta y(t)$ and $\Delta z(t)$, and rotations around the axes $\Delta \theta_{x}(t)$, $\Delta \theta_{y}(t)$ and $\Delta \theta_{z}(t)$. Due to the assumption of a distant light source, $\Delta x(t)$, $\Delta y(t)$ and $\Delta \theta_{z}(t)$, which are orthogonal to $\vec{l}$, will have no influence on $I_p(t)$. With interferences like talking and eating avoided, the skin deformation $\Delta r(t)$ caused by carotid pulse will be the only non-rigid motion.

Next, we elaborate on how the motions $\Delta z(t)$, $\Delta \theta_{x}(t)$, $\Delta \theta_{y}(t)$ and $\Delta r(t)$ influence the illumination of the neck point $I_p(t)$. First, the relationship between the light intensity $I_i(t)$ and the camera distance $z(t)$ obeys the inverse-square law:
\begin{equation}
	I_i(t)=\frac{I_0}{a+bz(t)+cz^2(t)},~z(t)=z_p+\Delta z(t)
\end{equation}
in which $I_0$ is the light source intensity before distance attenuation, $z_p$ is the distance of point $p$ to the camera, and $a$, $b$ and $c$ are constants. To simplify the notation, assume $B(z)=I_0/(a+bz+cz^2)$. Since $\Delta z(t) \ll z_p$, we can rewrite (2) as
\begin{equation}
	I_i(t)=B(z(t))=B(z_p+\Delta z(t))\approx B(z_p)+B'(z_p)\Delta z(t)
\end{equation}
Second, the strengths of both the diffuse reflection and the specular reflection are dependent on the direction of the surface normal $\vec{n}$. In Fig. \ref{phong} (b), $\vec{n}_{zx}$ is the projection of $\vec{n}$ in the z-x plane, and $\vec{n}_{yz}$ is the projection of $\vec{n}$ in the y-z plane. Assuming $\theta_y(t)$ to be the angle between $\vec{n}_{zx}$ and the $z$ axis, and $\theta_x(t)$ to be the angle between $\vec{n}_{yz}$ and the $z$ axis, we can get
\begin{equation}
	\vec{l}\cdot\vec{n}(t)=\vec{n}(t)\cdot\vec{h}=\frac{1}{\sqrt{1+\tan^2\theta_x(t)+\tan^2\theta_y(t)}}
\end{equation}
Both $\theta_x(t)$ and $\theta_y(t)$ can be separated into a stationary part and a varying part:
\begin{equation}
	\theta_x(t) = \theta_x + \Delta\theta_x(t)
\end{equation}
\begin{equation}
	\theta_y(t) \approx \left\{
	\begin{array}{lr}
		\theta_y + \Delta\theta_y(t), & \theta_y<\theta_0-\sqrt{\frac{2\Delta r(t)}{r}}\\
		\theta_y + \Delta\theta_y(t)-k\Delta r(t), & \theta_y\geq\theta_0-\sqrt{\frac{2\Delta r(t)}{r}}
	\end{array}
	\right.
\end{equation}
in which $k$ is a constant, $\theta_x(t)$ is affected by the rigid rotation around the $x$-axis, and $\theta_y(t)$ is affected by the rigid rotation around the $y$-axis and the non-rigid deformation (only when the point $p$ is close enough to the carotid arteries). To simplify the notations, assume $\Psi(\theta_x,\theta_y)=1/\sqrt{1+\tan^2\theta_x+\tan^2\theta_y}$ so that (4) can be rewritten as 
\begin{eqnarray}
	\vec{l}\cdot\vec{n}(t)=\vec{n}(t)\cdot\vec{h}=\Psi(\theta_x(t),\theta_y(t)) 
	\approx \Psi(\theta_x,\theta_y)+\frac{\partial\Psi}{\partial\theta_x}\Delta\theta_x(t)+\frac{\partial\Psi}{\partial\theta_y}(\Delta\theta_y(t)-k\Delta r(t)) 
\end{eqnarray}
Finally, by substituting (3) and (7) into (1), $I_p(t)$ becomes
\begin{eqnarray}
	I_p(t) &\approx I_a k_a + (B(z_p)+B'(z_p)\Delta z(t))\cdot(k_d\Psi(\theta_x,\theta_y)+\\ \nonumber
	&k_s\Psi^{\alpha}(\theta_x,\theta_y)+(k_d+k_s\alpha\Psi^{\alpha-1}(\theta_x,\theta_y))\cdot\\ \nonumber
	&(\frac{\partial\Psi}{\partial\theta_x}\Delta\theta_x(t)+\frac{\partial\Psi}{\partial\theta_y}\Delta\theta_y(t)-\frac{\partial\Psi}{\partial\theta_y}k\Delta r(t))) + I_n(t)
\end{eqnarray}
In (8), the camera quantization error $I_n(t)$ can be attenuated by spatially downsampling the video frames, and the product of the small varying terms can be neglected. Then we reach the final expression of $I_p(t)$
\begin{eqnarray}
	I_p(t) &\approx I_a k_a + k_dB(z_p)\Psi(\theta_x,\theta_y) + k_sB(z_p)\Psi^{\alpha}(\theta_x,\theta_y)\\ \nonumber
	&+(k_dB'(z_p)\Psi(\theta_x,\theta_y) + k_sB'(z_p)\Psi^{\alpha}(\theta_x,\theta_y))\cdot\Delta z(t)\\ \nonumber
	&+B(z_p)\frac{\partial\Psi}{\partial\theta_x}(k_d+k_s\alpha\Psi^{\alpha-1}(\theta_x,\theta_y))\cdot\Delta\theta_x(t)\\ \nonumber
	&+B(z_p)\frac{\partial\Psi}{\partial\theta_y}(k_d+k_s\alpha\Psi^{\alpha-1}(\theta_x,\theta_y))\cdot\Delta\theta_y(t)\\ \nonumber
	&-kB(z_p)\frac{\partial\Psi}{\partial\theta_y}(k_d+k_s\alpha\Psi^{\alpha-1}(\theta_x,\theta_y))\cdot\Delta r(t)\\ \nonumber
\end{eqnarray}
which is approximately a linear combination of $\Delta z(t)$ the distance change between the neck point and the camera, $\Delta\theta_x(t)$ the anterior-posterior rotation of the neck (e.g. nodding), $\Delta\theta_y(t)$ the neck rotation around the cervical spine, and $\Delta r(t)$ the skin deformation caused by the carotid pulse.

Among the four motion sources, $\Delta z(t)$ is usually much stronger than the others, so simply calculating the common average reference of $I_p(t)$ among all neck points $p\in[1,\cdots,N_p]$ will provide an estimate of $\Delta z(t)$:
\begin{equation}
	\Delta z(t) \propto \bar{I}_p(t)=\frac{1}{N_p}\sum_{p=1}^{N_p}I_p(t)
\end{equation}
Most human actions can affect $\Delta z(t)$. When there is no other large motion, $\Delta z(t)$ will be dominated by breathing movements, and sometimes also contain a weak ballistocardiography signal. The breathing signal and the pulse-induced ballistocardiography signal can be further separated in the frequency domain.

To extract the carotid pulse deformation $\Delta r(t)$ from $I_p(t)$, the strong $\Delta z(t)$ needs to be attenuated. Since $\Delta z(t)$ is a global translation homogeneous on almost all neck points whereas $\Delta r(t)$ only exists on points satisfying $\theta_y<\theta_0-\sqrt{2\Delta r(t)/r}$, calculating $I_p(t)-\bar{I}_p(t)$ will diminish $\Delta z(t)$ while preserving $\Delta r(t)$. In the remaining signals, $\Delta r(t)$ is uncorrelated with the other motion sources $\Delta\theta_y(t)$ and $\Delta\theta_x(t)$, so we can use Principal Component Analysis~(PCA)~\cite{jolliffe2002principal} to separate it. In the PCA transformed domain, the first three principal components of $I_p(t)-\bar{I}_p(t)$ with the highest variances will correspond to $\Delta r(t)$, $\Delta\theta_y(t)$ and $\Delta\theta_x(t)$, but their order is not guaranteed.


In brief, the breathing movement can be estimated from $\bar{I}_p(t)$, while the pulse signal can be selected from $\bar{I}_p(t)$ (when ballistocardiography exists in it) and the first three principal components of $I_p(t)-\bar{I}_p(t)$.

\begin{figure}[!th]
	\centering\includegraphics[width=0.8\linewidth]{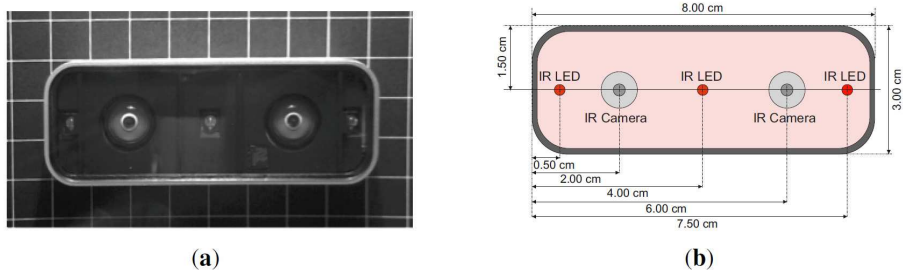}
	\caption{Visualization of a real (a) and schematic view (b) of the Leap Motion controller \cite{Weichert2013}. Note that the optical filter of the device was removed in (a) only to reveal its inside. It was kept intact in our experiments.}
	\label{leap}
\end{figure}

\section{Apparatus}


A large variety of NIR imaging devices are available in the market. This work uses the Leap Motion controller, which is a computer hardware sensor device that supports non-contact hand and finger interactions, analogous to a mouse. Using two monochromatic infrared cameras and three infrared LEDs (Fig. \ref{leap}), the device observes a roughly hemispherical area, to a distance of about 1 m. The LEDs generate NIR light at a wavelength of 850 nanometers, and the cameras capture frames in 8 bits with a pixel resolution of 640 x 240, which are sent through a USB cable to the host computer at a floating frame rate around 62 frames per second (fps) and stored with timestamps. The recording program was written in Python using the Leap Motion SDK, while all the data analysis was done in MATLAB R2015b (MathWorks, Inc., Natick, MA, USA). To make the proposed methods transferable to devices with a single NIR camera, only the frames from one camera were used.   

\section{Experimental Protocol}
Twelve participants (8 males, 4 females) between the ages of 23-34 years were recruited for this study, which was pre-approved by the Massachusetts Institute of Technology review board. The participant sample covers both genders, different ages, and varying skin colors (Asians, Africans and Caucasians). Informed consent was obtained from all the participants prior to each study session.

\begin{figure}[!th]
	\centering\includegraphics[width=0.45\linewidth]{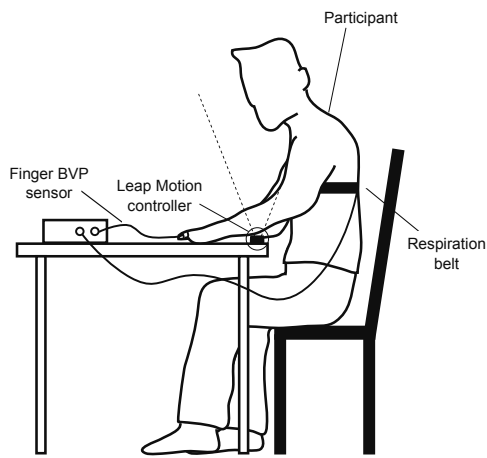}
	\caption{Experimental setup (BVP: blood volume pulse).}
	\label{setup}
\end{figure}

\begin{figure}[!th]
	\centering\includegraphics[width=0.95\linewidth]{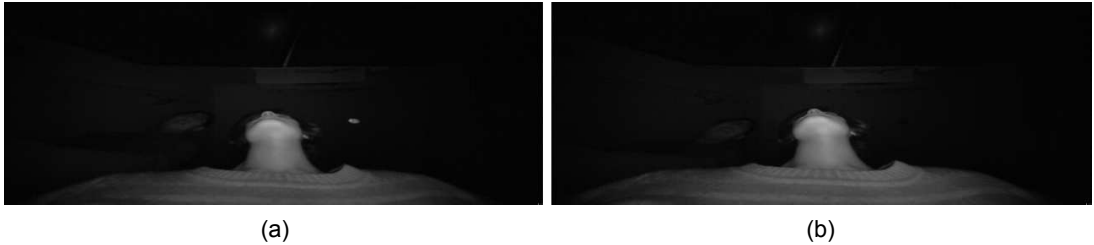}
	\caption{Infrared frames captured by one of the Leap Motion cameras in a bright room (a) and a fully dark room (b).}
	\label{brightdark}
\end{figure}

In each experiment, participants were seated at a desk naturally (Fig. \ref{setup}). The Leap Motion controller was placed on the edge of the desk, parallel with it, and facing upward. Due to different heights and postures of participants, the distance between the controller cameras and the neck of participants varied. Gold standard physiological measurements were obtained with an FDA-cleared sensor (FlexComp Infiniti by Thought Technologies~Ltd.) that simultaneously recorded Blood Volume Pulse (BVP) from a finger probe and breathing waveform (BW) from a chest belt at a constant sampling frequency of 256~Hz. The sensor data were synchronized with the NIR frames via timestamps. Every experiment consists of two phases. In the first phase, the room was bright (184 lux at the camera place), while in the second phase all the lights were turned off to make the room fully dark (1 lux at the camera place). Participants were asked to breathe spontaneously and avoid major movement for 1 min during each of the phases. Fig. \ref{brightdark} shows that there is no visible difference in image brightness and quality between infrared frames captured in a bright room and a fully dark room by the NIR cameras.

\begin{figure}[!b]
	\centering\includegraphics[width=\linewidth]{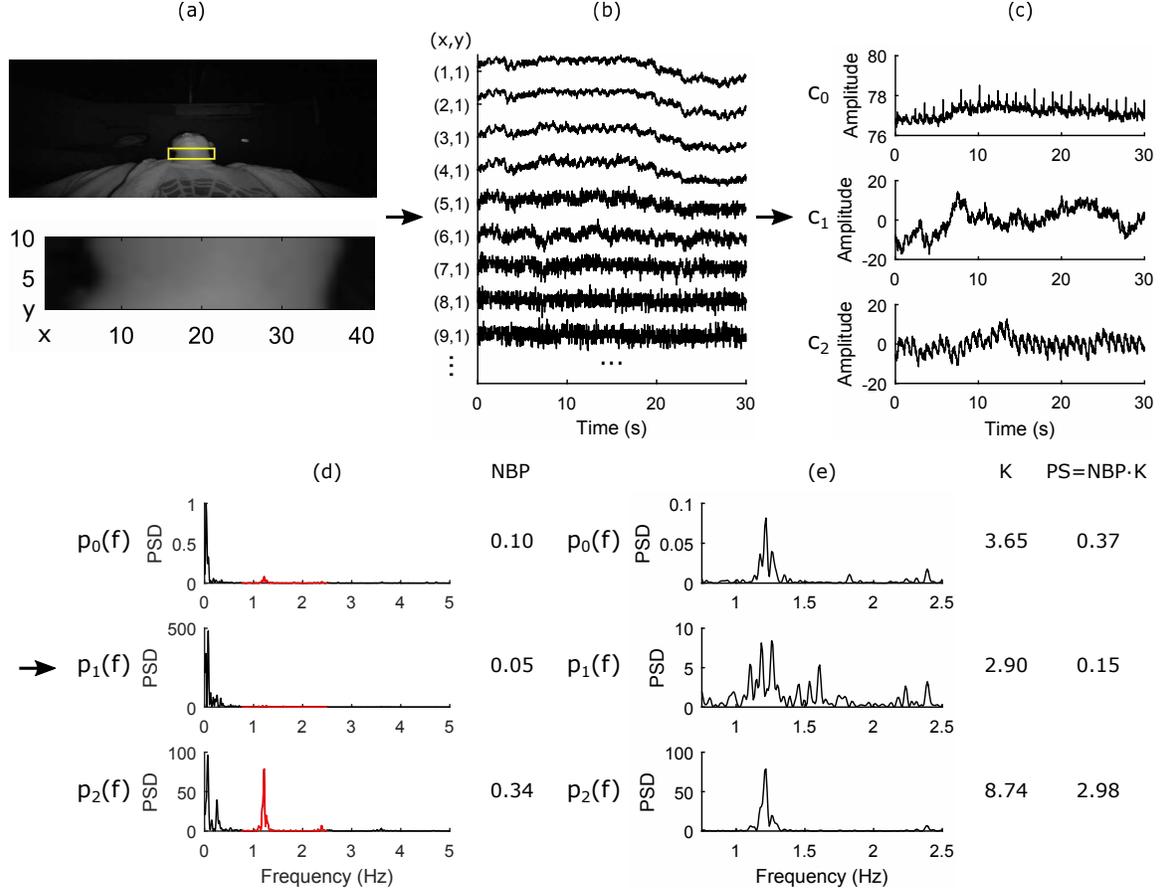}
	\caption{Flow chart for HR measurement. (a) An exemplary NIR frame with the detected 81~by~19~pixels neck area marked in yellow, and the enlarged version of the area after being downsampled to 41 pixels x 10 pixels. (b) Time series showing the intensity changes of the pixels on the neck area. (c) Component candidates: $c_0$ the common average of all the time series, $c_1$ the second principal component, and $c_2$ the third principal component. (d) PSD estimates of the three component candidates within [0 Hz, 5 Hz], and their normalized band power (NBP) (e) The same PSD estimates within the prospective HR frequency band [0.75 Hz, 2.5 Hz], their kurtosis (K) and pulse significance (PS).}
	\label{flowHR}
\end{figure}

\section{Heart Rate Estimation}

This section describes the extraction of heart rate from carotid pulse movements which is depicted in Fig. \ref{flowHR}.

\subsection{Region-of-interest Detection}

\begin{figure}[!th]
	\centering\includegraphics[width=0.5\linewidth]{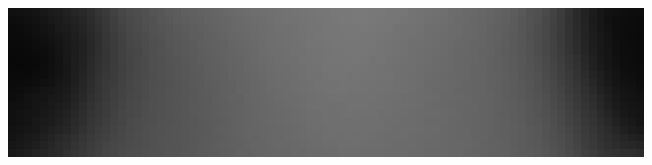}
	\caption{Template image of the neck area used in template matching.}
	\label{template}
\end{figure}

\begin{figure}[!th]
	\centering\includegraphics[width=0.95\linewidth]{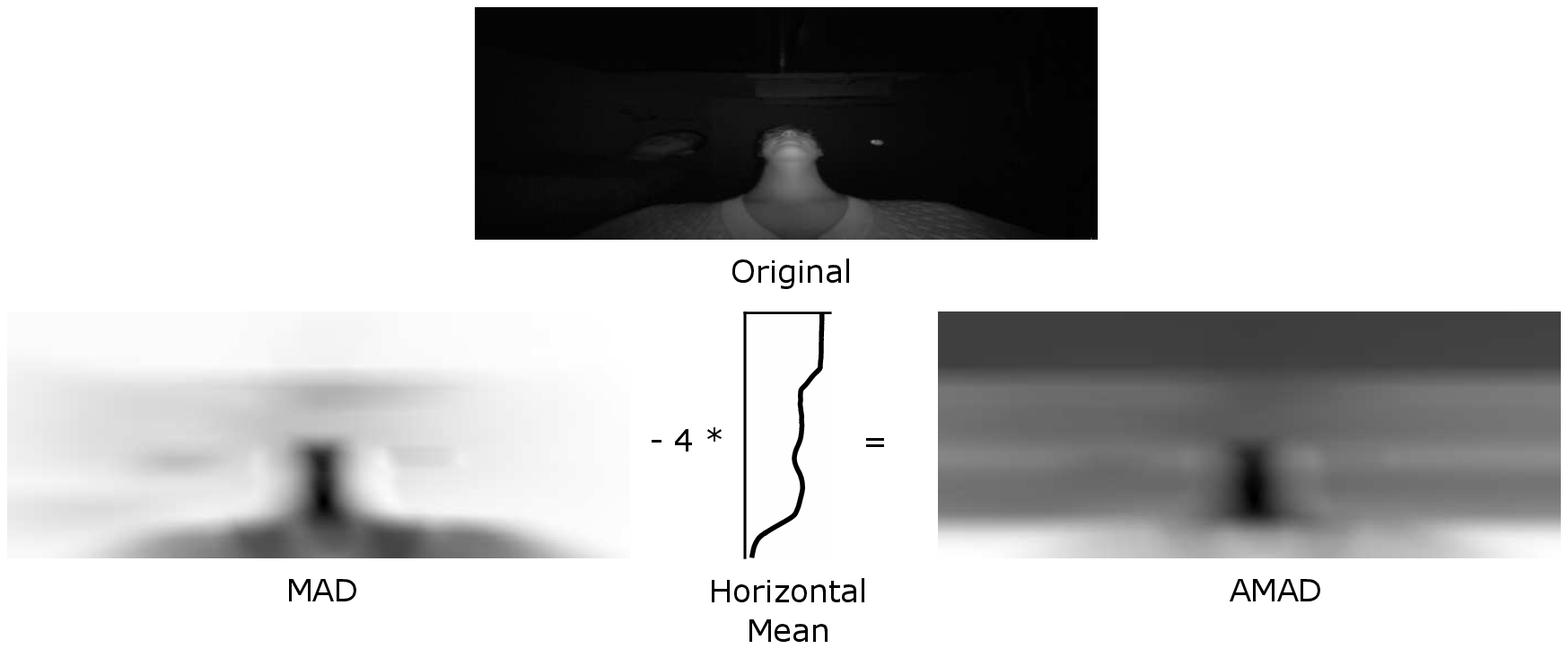}
	\caption{A NIR frame, and how its AMAD map was derived from the MAD map in neck detection.}
	\label{AMAD}
\end{figure}

The neck area (a rectangular area including the neck) was automatically located in every NIR video using template matching \cite{brunelli2009template}. In particular, a template (Fig. \ref{template}) was created by manually cropping neck areas from 48 static infrared photos of the 12 participants with various postures and lighting conditions, resizing them to the same size (81~by~19~pixels) and taking the average of them.  For each video, the template matching algorithm slides the template image over the first frame, and compares the template and patch of the frame by calculating an adjusted mean of absolute differences (AMAD). As shown in Fig. \ref{AMAD}, a mean of absolute differences (MAD) metric averages the absolute values of the differences between pixels in the original frame and the corresponding pixels in the template image. To enhance the contrast between the neck and the other body parts, the geometric feature of the neck was exploited. Since in a MAD map the neck is always horizontally the slimmest part compared with the head and the torso, the horizontal mean was calculated and subtracted from every column of the map after multiplication with a factor of 4 (chosen heuristically to balance the neck-torso contrast against the neck-background contrast) to generate the AMAD map:
\begin{equation}
	MAD(x,y) = \frac{1}{T_{rows}T_{cols}}\sum_{i=0}^{T_{rows}-1}\sum_{j=0}^{T_{cols}-1}|I_s(x+i,y+j)-I_t(i,j)|
\end{equation}
\begin{equation}
	AMAD(x,y) = MAD(x,y) - \frac{4}{S_{cols}-T_{cols}+1}\sum_{y=0}^{S_{cols}-T_{cols}}MAD(x,y)
\end{equation}
in which $x=0,\cdots,S_{rows}-T_{rows},~y=0,\cdots,S_{cols}-T_{cols}$, $I_s$ is the search image with a dimension of $S_{rows}\times S_{cols}$, and $I_t$ is the template with a dimension of $T_{rows}\times T_{cols}$ ($T_{rows}\leq S_{rows}$ and $T_{cols}\leq S_{cols}$).

As described in the experimental setup, there was no restriction on the distance between the camera and the neck of a participant, which caused the sizes of the necks to vary in videos. To make the neck detection more scale-robust, the template was resized by a scale of 0.8 (based on the smallest neck area we observed among the static photos), and AMAD maps were computed twice using both the original template and the resized one. Finally, the lowest AMAD score among the two maps indicated the optimal size of the template and its best position within the frame. The size and the position were then applied to all the frames of the video for neck segmentation.

\subsection{Spatial Denoising}

To reduce random noise produced by the camera and the sensor circuits, the neck area segmented in every frame was downsampled to half of its size using bicubic interpolation, which helps reduce the computational cost of subsequent operations.

\subsection{Frequency Denoising}
The intensity change of every pixel within the downsampled neck area formed a 1-min time series, which can be considered as a channel of sensor recording. However, a single channel usually exhibited a mixture of the carotid pulse vibration, breathing movement and motion artifacts. To separate the different signals, we used PCA to combine all the channels (all the pixels within the neck area) and convert these correlated time series into a set of linearly uncorrelated components. The component corresponding to the pulse signal was then automatically selected from them based on features in the frequency domain. The procedure is illustrated in Fig. \ref{flowHR} with details given below:

\begin{enumerate}
	\item A moving window of 30 s in length with $96.7\%$ overlap (1 s increment) was applied to every 1-min time series.
	\item Within each window, a common average reference (CAR) montage was performed by subtracting to each channel the average of all channels. The common average related to the global movement of the neck was saved as a component candidate $c_0$, which was dominated by breathing movement in most cases but sometimes also showed weak ballistocardiography.
	\item PCA was applied on all the channels with the common average removed to return principal component scores in descending order of component variance. According to observation on the three scores with the highest variances, the first component was usually associated with motion artifacts, and the carotid pulse was visible in the second, the third or both depending on the posture of the participant and the distance between the camera and the neck. Therefore, the second and the third principal components were saved as component candidates $c_1$ and $c_2$.
	\item To select the component best representing the pulse, the power spectral density (PSD) estimates $p_0(f)$, $p_1(f)$ and $p_2(f)$, $f\in (0,f_N)$ ($f_N$ is the Nyquist frequency), were computed from $c_0$, $c_1$ and $c_2$ using the Lomb-Scargle method \cite{lomb1976least,scargle1982}, which is good at finding weak periodic signals in otherwise random unevenly sampled data. The method was chosen over other PSD estimation methods, mainly because the Leap Motion controller captures NIR video at a non-constant sampling rate.
	\item A single distinct peak in the PSD estimate represents a well-detected HR. However, with the existence of noise, a HR peak might be in company with other peaks and may not be significantly higher and/or sharper. To evaluate the distinction of cardiac pulse in a PSD estimate, a measure called pulse significance (PS) was invented by combining two metrics, the normalized band power (NBP) and the kurtosis. The NBP reflects how high a HR peak is in a PSD estimate. It was calculated as the power within the prospective HR frequency band [0.75~Hz, 2.5~Hz] (corresponding to 45 and 150 beats per minute) divided by the full spectrum power:
	\begin{equation}
		NBP(p_k) = \frac{\sum_{f=0.75}^{2.5}p_k(f)}{\sum_{f=0}^{f_N}p_k(f)},~k=0,1,2
	\end{equation}
	The kurtosis, also called the fourth standardized moment, indicates the primary peakedness and lack of shoulder in a distribution \cite{decarlo1997meaning}, which in our method measures how sharp a peak is within the HR frequency band. It was derived as
	\begin{equation}
		\mu_k = \frac{\sum_{f=0.75}^{2.5}p_k(f)f}{\sum_{f=0.75}^{2.5}f},~k=0,1,2
	\end{equation}		
	\begin{equation}
		K(p_k) = \frac{E[(p_k(f)-\mu_k)^4]}{(E[(p_k(f)-\mu_k)^2])^2} = \frac{[\sum_{f=0.75}^{2.5}(p_k(f)-\mu_k)^4f][\sum_{f=0.75}^{2.5}f]}{[\sum_{f=0.75}^{2.5}(p_k(f)-\mu_k)^2f]^2},~k=0,1,2
	\end{equation}	
	Finally, the NBP and the kurtosis were multiplied with each other to get the pulse significance:
	\begin{equation}
		PS(p_k) = NBP(p_k)\cdot K(p_k),~k=0,1,2
	\end{equation}	
	The component among $c_0$, $c_1$ and $c_2$ whose PSD estimate had the largest PS was selected for HR extraction, and denoted as $c$.
\end{enumerate}

\begin{figure}[!th]
	\centering\includegraphics[width=0.4\linewidth]{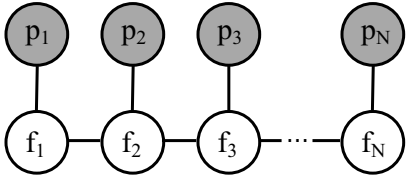}
	\caption{The Hidden Markov Model designed for HR smoothing. The gray nodes were observed PSD estimates, and the white nodes were the heart rates we wished to infer.}
	\label{hmm}
\end{figure}

\begin{figure}[!th]
	\centering\includegraphics[width=\linewidth]{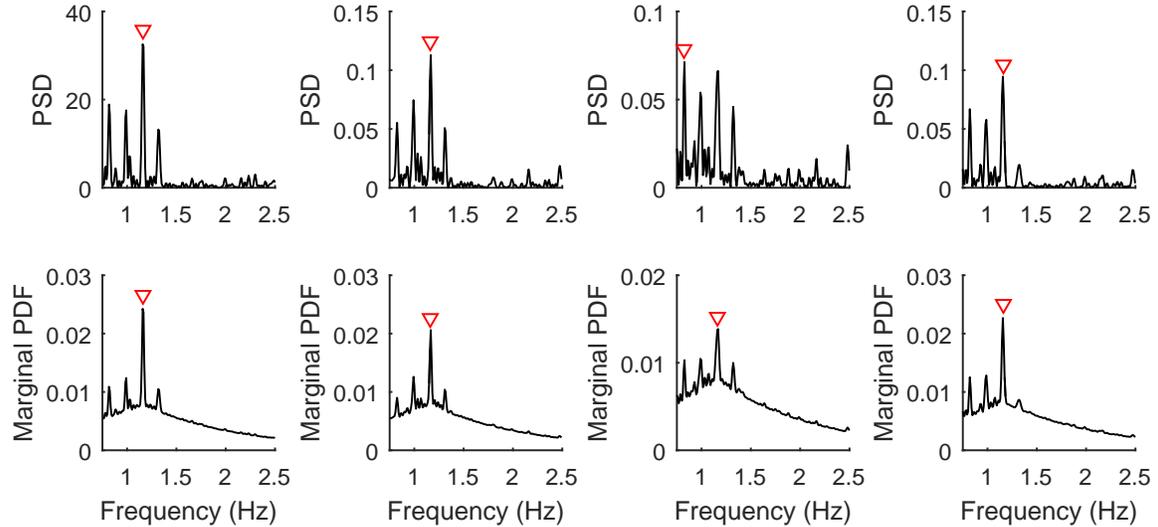}
	\caption{PSD estimates $p_i(f)$ and marginal distributions within the prospective HR frequency band for four consecutive windows. The red triangles denote the maxima of each spectrum.}
	\label{margs}
\end{figure}

\subsection{Temporal Denoising}


After the optimal component $c$ corresponding to the carotid pulse was selected, the HR could be easily read from its PSD estimate $p(f)$ by locating the highest peak within the HR frequency band. However, sometimes the motion artifact was so strong that the HR reading of a window significantly deviated from its neighbors, which violated the continuity of the human heart rate. To address the problem, we designed a Hidden Markov Model (HMM, Fig. \ref{hmm}) to smooth readings in consecutive windows and used belief propagation to infer the HR. Each pair of gray and white nodes in the Markov chain corresponded to a time window. The gray nodes were observed PSD estimates, and the white nodes were the heart rates we wished to infer. Each window was connected to only its neighbors, and favored a smooth HR variance. Mathematically, we can write this chain with the probability density function (PDF):
\begin{equation}
	P(f|p(f)) = \frac{1}{Z}\phi(f_1|p_1(f))\prod_{i=2}^{N}\phi(f_i|p_i(f))\psi(f_i,f_{i-1})
\end{equation}
where $f_i\in[0.75, 2.5]Hz$ is the HR frequency of window i, which we wished to infer given $p(f)$. $Z$ is a normalization factor to guarantee the integral of the PDF equals 1. $\phi(f_i|p_i(f))$ is known as the data term and $\psi(f_i,f_{i-1})$ is know as the smoothness terms. We defined them to be:
\begin{equation}
	\phi(f_i|p_i(f))\propto\exp(\lambda\frac{p_i(f_i)}{\sum_{f=0.75}^{2.5}p_i(f)})
\end{equation}
\begin{equation}
	\psi(f_i,f_{i-1})\propto\exp(-|f_i-f_{i-1}|)
\end{equation}
in which the parameter $\lambda\in\mathbb{R}$ allows control over the relative strength of the data term versus the smoothness term. We used $\lambda = 16$ in this study to match the normal variation of physiological parameters under resting state. In use cases where the HR can change more abruptly, $\lambda$ should be increased.

Belief propagation was used to obtain the maximum a posteriori (MAP) estimation of the possible HR frequencies by passing messages left-to-right in the chain. The message update equation is
\begin{equation}
	m_{p\to q}(f_q)=\max_{f_p}\phi(f_p)\psi(f_p,f_q)\prod_{s\in N(p)\backslash q}m_{s\to p}(f)
\end{equation}
where the expression $s\in N(p)\backslash q$ is all the neighbors of node $p$ except node $q$. Finally, in the marginal distribution of each window, the HR was read by finding the maximum. Fig. \ref{margs} shows an example of how HMM smoothing corrects a deviated HR data point.

\section{Breathing Rate Estimation}

\begin{figure}[!th]
	\centering\includegraphics[width=\linewidth]{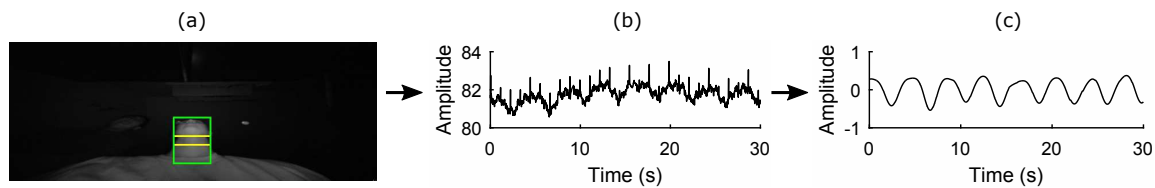}
	\caption{Flow chart for BR measurement. (a) An exemplary NIR frame with the detected neck area marked in yellow and the respiration ROI marked in green. (b) The spatial average of all the pixel intensities within the ROI, fluctuating with time. (c) The breathing movement estimated by band-pass filtering the time series in (b).}
	\label{flowRR}
\end{figure}

This section describes the extraction of breathing rate from movements of the neck which is depicted in Fig. \ref{flowRR}.

\subsection{Region-of-interest Detection}
While the carotid pulse is reflected as local motions on the two sides of the neck, the breathing activity is associated with the global movement of the neck and the torso. To extract the BR, the same neck detection algorithm we used in HR measurement was applied. However, the region-of-interest  was expanded to 5 times its height (twice above and twice below) to cover a complete neck area as well as the chin and a small part of the upper chest.

\subsection{Spatial Denoising}
To improve the signal-to-noise ratio (SNR), the intensities of all pixels within the region-of-interest (ROI) were averaged in every frame to form a single 1-min time series for each video. The same moving window (30 s in length and 1 s in increment) used for HR measurement was employed on the series.

\subsection{Frequency Denoising}
After spatial averaging, most local motions like the carotid pulse are neutralized, but some global motion artifacts might still exist. To remove noise, a 3rd order band-pass Butterworth filter with cut-off frequencies of 0.08 and 0.5 Hz (corresponding to 4.8 and 30 breaths per minute) was applied to each windowed signal. To maintain synchrony with the gold standard data, zero-phase filtering was performed by processing the signal in both the forward and reverse directions \cite{oppenheim1989discrete}.

\subsection{Temporal Denoising}
The Lomb-Scargle method was used again to obtain the PSD estimate of the filtered signal. The same Hidden Markov Model as in Fig. \ref{hmm} was applied to consecutive windows of PSD estimates, and BR was selected to be the frequency where the marginal distribution achieved the maximum in every window.

\section{Results}
This section describes the performance obtained for heart rate and breathing rate estimation from the neck when compared with an FDA-cleared device. 

\subsection{Heart Rate}

\begin{figure}[!th]
	\centering\includegraphics[width=\linewidth]{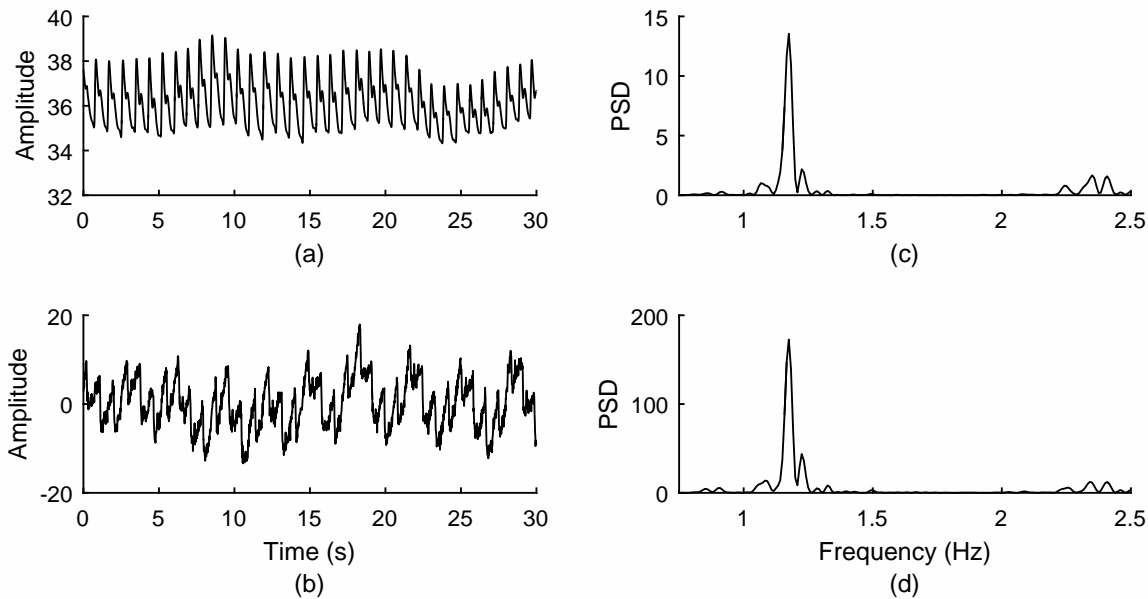}
	\caption{An exemplary 30-second window in HR measurement. (a) Blood volume pulse waveform measured by the finger probe. (b) The optimal component $c$ extracted from the neck corresponding to the carotid pulse. (c) Lomb-Scargle periodogram of the BVP signal within the prospective HR frequency band. (d) Lomb-Scargle periodogram of $c$ within the prospective HR frequency band.}
	\label{compHR}
\end{figure}

The time series and frequency spectra of the cardiac pulse measured by our proposed methodology and the reference device are compared in an exemplary window in Fig. \ref{compHR}. Though the time series shares the same periodicity, its wave patterns in each period look different. On one hand, this is because the carotid pulse observed over the skin surface can not faithfully reflect the blood volume changes inside the subjacent arteries. On the other hand, the reference BVP exhibits more details because of its higher sampling rate (256 Hz) compared with the NIR camera (around 62 Hz). Nevertheless, in the frequency domain within the prospective HR frequency band [0.75 Hz, 2.5 Hz], the periodograms of the two pulse signals look very similar to each other, including both the fundamental tone (around 1.17 Hz) and the 2nd harmonic (around 2.34 Hz) of the HR.

\begin{figure}[!th]
	\centering\includegraphics[width=\linewidth]{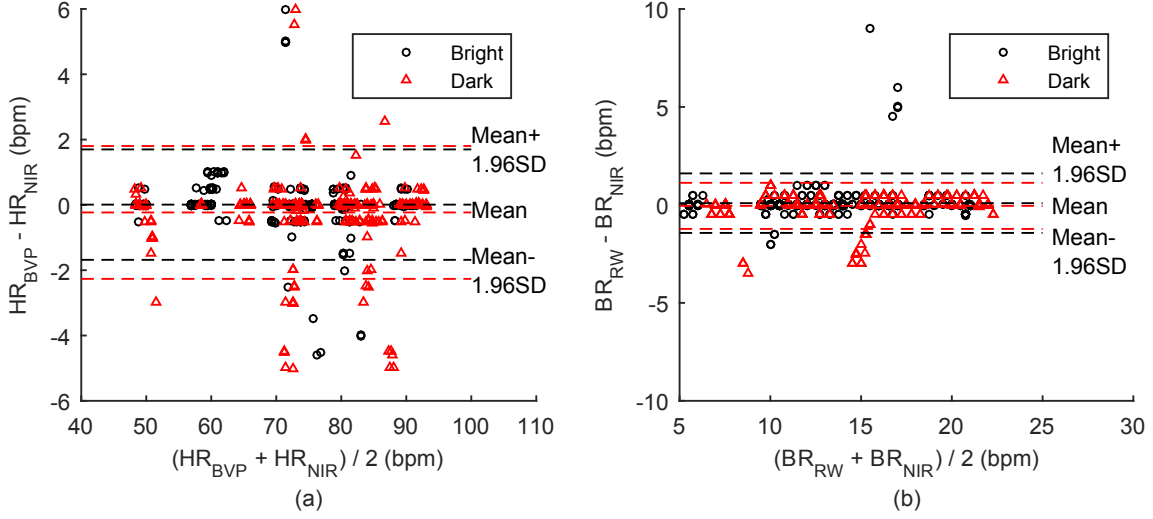}
	\caption{Bland-Altman plots for overall 744 pairs of (a) HR measurements and (b) BR measurements, with two halves of these pairs respectively measured in bright and dark conditions (HR: heart rate, BVP: blood volume pulse, BR: breathing rate, BW: breathing waveform, NIR: near-infrared, bpm: beats/breaths per minute).}
	\label{ba}
\end{figure}

The agreement between the 744 pairs of HR measurements from 12 participants was tested by Bland-Altman analysis \cite{bland1986statistical}. According to Fig. \ref{ba} (a), our HR measurements ranged from 48.5 bpm to 93.0 bpm, which almost cover the normal resting adult HR range 60-100 bpm. No trend of increased measurement error can be observed from both  low and high ends of the range. With measurements under bright and dark conditions analyzed separately, their Bland-Altman plots, superimposed in the same figure, have similar distributions.

Table \ref{statHR} summarizes the descriptive statistics for comparing the proposed methodology with the gold standard. The overall mean absolute error (MAE) between them was 0.36 bpm, which proved the high accuracy of our technique. To verify the robustness of our NIR-based method under different lighting environments, a paired t-test was conducted to compare the absolute errors of HR measurements in bright and dark conditions. No  significant differences were found in terms of error during bright (M = 0.31, SD = 0.81) and dark (M = 0.41, SD = 0.98) conditions; t(371) = -1.572, p = 0.117.

\begin{figure}[!th]
	\centering\includegraphics[width=\linewidth]{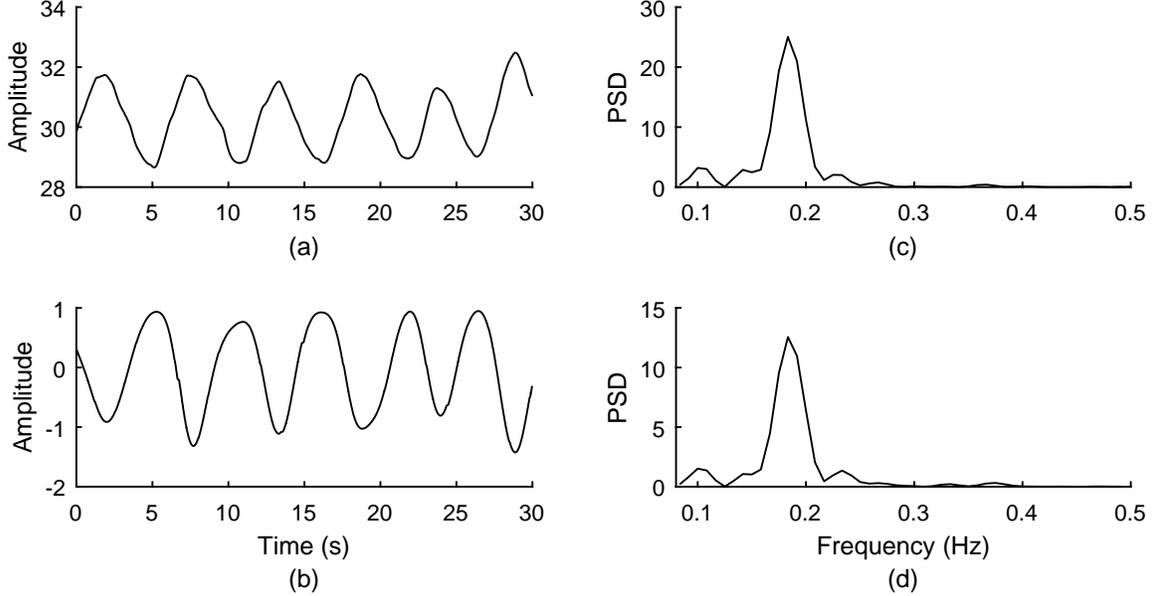}
	\caption{An exemplary 30-second window in BR measurement. (a) Breathing waveform measured by the respiration belt. (b) Breathing movement estimated from the NIR video. (c) Lomb-Scargle periodogram of the breathing waveform within the prospective BR frequency band. (d) Lomb-Scargle periodogram of the estimated breathing movement within the prospective BR frequency band.}
	\label{compRR}
\end{figure}


\begin{table}[htbp]
	\centering\caption{\label{statHR}Heart rate measurement}
	\begin{tabular}{@{}llll}
			\hline
			Statistic                 & Overall & Bright & Dark  \\
			\hline
			No. of measurement pairs  & 744     & 372    & 372   \\
			Mean absolute error (bpm) & 0.36    & 0.31   & 0.41  \\
			Mean error (bpm)          & -0.11   & 0.01   & -0.23 \\
			SD of error (bpm)         & 0.96    & 0.86   & 1.04  \\
			Root-mean-square error    & 0.97    & 0.86   & 1.06  \\
			Correlation coefficient   & 1.00    & 1.00   & 1.00  \\
			\hline
	\end{tabular}	
\end{table}


\begin{table}[htbp]
	\centering\caption{\label{statRR}Breathing rate measurement}
	\begin{tabular}{@{}llll}
			\hline
			Statistic                 & Overall & Bright & Dark  \\
			\hline
			No. of measurement pairs  & 744     & 372    & 372   \\
			Mean absolute error (bpm) & 0.24    & 0.22   & 0.26  \\
			Mean error (bpm)          & 0.02    & 0.09   & -0.05 \\
			SD of error (bpm)         & 0.70    & 0.77   & 0.60  \\
			Root-mean-square error    & 0.70    & 0.78   & 0.60  \\
			Correlation coefficient   & 0.99    & 0.98   & 0.99  \\
			\hline
		\end{tabular}	
\end{table}

\subsection{Breathing Rate}

Within a 30-second window, we plotted the time series and frequency spectra of the breathing movement measured by our proposed methodology and the reference device for comparison (Fig. \ref{compRR}). Both the pair of time series and the pair of periodograms within the prospective BR frequency band [0.08 Hz, 0.5 Hz] show high similarity in shape, though the two time series have opposite phases.

A Bland-Altman plot was drawn for all BR measurement pairs in Fig. \ref{ba} (b). The measurements covered a BR range between 5.0 and 22.0 breaths per minute, where our method manifested high accuracy consistently except only a few outliers due to severe motion artifacts. The figure marks data points under two different lighting conditions differently, of which both the means and the standard deviations (SD) are alike.

The quantitative statistics for comparing the proposed methodology with the gold standard are listed in Table \ref{statRR}. The overall MAE of BR between the two methods was 0.24 bpm, even smaller than the same statistic for HR. To verify the robustness of our NIR-based method under different lighting environments, a paired t-test was conducted to compare the absolute errors of BR measurements in bright and dark conditions. Similarly to HR, no significant difference was found in terms of error during bright (M~=~0.22,~SD~=~0.75) and dark (M~=~0.26, SD~=~0.54) conditions; t(371)~=~-0.936,~p~=~0.350.


\begin{table}[!t]
	\centering
	\begin{threeparttable}
		\caption{\label{stat}Statistics with respect to each video$^a$}
		\footnotesize		
		\begin{tabular}{@{}lccrrrcc}
			\hline
			Participant $\#$ & Gender & ROI size & $c_0~\%$ & $c_1~\%$ & $c_2~\%$ & $MAE_{HR}$ & $MAE_{BR}$ \\
			\hline
			1 (Bright) & M & 81x19 & $0.00\%$ & $22.58\%$ & $77.42\%$ & 0.02 & 1.15 \\
			1 (Dark) & M & 81x19 & $0.00\%$ & $96.77\%$ & $3.23\%$ & 0.09 & 0.13 \\
			2 (Bright) & F & 65x16 & $0.00\%$ & $29.03\%$ & $70.97\%$ & 1.36 & 0.11 \\
			2 (Dark) & F & 65x16 & $0.00\%$ & $54.84\%$ & $45.16\%$ & 2.01 & 0.52 \\
			3 (Bright) & M & 65x16 & $12.90\%$ & $6.45\%$ & $80.65\%$ & 0.07 & 0.02 \\
			3 (Dark) & M & 65x16 & $93.55\%$ & $0.00\%$ & $6.45\%$ & 0.37 & 0.08 \\
			4 (Bright) & F & 81x19 & $54.84\%$ & $6.45\%$ & $38.71\%$ & 0.65 & 0.59 \\
			4 (Dark) & F & 81x19 & $16.13\%$ & $67.74\%$ & $16.13\%$ & 0.09 & 0.39 \\
			5 (Bright) & M & 65x16 & $0.00\%$ & $32.26\%$ & $67.74\%$ & 0.06 & 0.10 \\
			5 (Dark) & M & 65x16 & $0.00\%$ & $45.16\%$ & $54.84\%$ & 0.50 & 0.05 \\
			6 (Bright) & F & 65x16 & $100.00\%$ & $0.00\%$ & $0.00\%$ & 0.44 & 0.07 \\
			6 (Dark) & F & 65x16 & $80.65\%$ & $0.00\%$ & $19.35\%$ & 1.41 & 1.05 \\
			7 (Bright) & M & 81x19 & $0.00\%$ & $80.65\%$ & $19.35\%$ & 0.11 & 0.03 \\
			7 (Dark) & M & 81x19 & $0.00\%$ & $16.13\%$ & $83.87\%$ & 0.12 & 0.18 \\
			8 (Bright) & M & 81x19 & $0.00\%$ & $22.58\%$ & $77.42\%$ & 0.03 & 0.08 \\
			8 (Dark) & M & 81x19 & $0.00\%$ & $9.68\%$ & $90.32\%$ & 0.09 & 0.15 \\
			9 (Bright) & M & 81x19 & $0.00\%$ & $0.00\%$ & $100.00\%$ & 0.04 & 0.04 \\
			9 (Dark) & M & 81x19 & $0.00\%$ & $9.68\%$ & $90.32\%$ & 0.01 & 0.13 \\
			10 (Bright) & M & 81x19 & $9.68\%$ & $0.00\%$ & $90.32\%$ & 0.14 & 0.03 \\
			10 (Dark) & M & 81x19 & $0.00\%$ & $0.00\%$ & $100.00\%$ & 0.07 & 0.02 \\
			11 (Bright) & F & 81x19 & $35.48\%$ & $0.00\%$ & $64.52\%$ & 0.24 & 0.25 \\
			11 (Dark) & F & 81x19 & $3.23\%$ & $0.00\%$ & $96.77\%$ & 0.06 & 0.20 \\
			12 (Bright) & M & 81x19 & $67.74\%$ & $12.90\%$ & $19.35\%$ & 0.54 & 0.15 \\
			12 (Dark) & M & 81x19 & $74.19\%$ & $0.00\%$ & $25.81\%$ & 0.10 & 0.26 \\
			\hline
			r with $MAE_{HR}$ & 0.59 & -0.58 & 0.21 & 0.09 & -0.28 &  &  \\
			p-value & 0.003 & 0.003 & 0.322 & 0.689 & 0.180 &  &  \\
			\hline
			r with $MAE_{BR}$ & 0.37 & -0.02 &  &  &  &  &  \\
			p-value & 0.070 & 0.922 &  &  &  &  &  \\
			\hline
		\end{tabular}
		\begin{tablenotes}
			\item[] $^a$ROI: region-of-interest, MAE: mean absolute error, HR: heart rate, BR: breathing rate, M: male, F: female.
		\end{tablenotes}
	\end{threeparttable}			
\end{table}
\normalsize

\section{Discussion}

This work demonstrates that low-cost NIR imaging devices can be used to provide accurate heart rate and breathing rate estimations from neck motions in well-lit and dark settings. While the results are encouraging for the data we collected, there are several factors that need to be considered when assessing the generalization of our findings.

Table \ref{stat} shows a significant positive point-biserial correlation between gender and $MAE_{HR}$ values ($r=0.59, p=0.003$), leading to lower accuracy for female participants. This difference may be partly due to the smaller size of carotid arteries of females~\cite{Krejza2006} which causes the carotid pulse to be subtler too. Table \ref{stat} also shows a significant negative point-biserial correlation between the  ROI sizes and their $MAE_{HR}$ values ($r=-0.58, p=0.003$), leading to lower accuracy for smaller ROI sizes. 
This difference was partly due to the different heights and/or tendency for some people to naturally bury their heads when working at their desk. In contrast, no significant correlations were observed when considering the relationship between $MAE_{BR}$ values and gender ($r=-0.37, p=0.070$) and ROI size ($r=-0.02, p=0.922$). These findings are partly to be expected as breathing movement has no gender-related distinctions, and the ROI size for BR was five times larger than the size of the ROI size for HR, making BR accuracy less sensitive to the variation of pixel numbers. Finally, no significant correlations were observed when considering the selected Principal Components (i.e.,~$c_0$, $c_1$, and $c_2$) and the $MAE_{HR}$.


\begin{figure}[!th]
	\centering\includegraphics[width=0.6\linewidth]{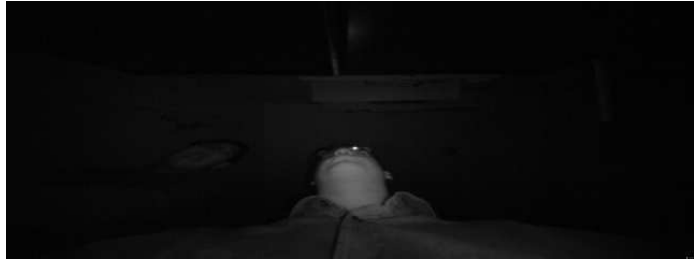}
	\caption{An example of a participant wearing a collared shirt.}
	\label{shirt}
\end{figure}

An obvious limiting factor of our proposed methodology is neck occlusions. While we do not expect our methods to work when the neck is not visible or fully covered (e.g.,~when wearing a scarf), our dataset contains several instances of partial neck occlusions. In particular, participants 3, 5 and 7 wore high collar shirts (e.g., Fig. \ref{shirt}), and participant 12 wore a turtleneck sweater. As shown in Table \ref{stat}, the maximum MAE values for these participants were 0.54 beats per minute and 0.26 breaths per minute, which demonstrates the effectiveness of the methods even with partial neck occlusions.

Finally, our study considered stationary conditions in which participants remained relatively still for few seconds. In particular, our approach makes two assumptions: 1)~the rigid motions of the neck are small enough so that the approximation in (9) will hold, and 2)~the non-rigid motions are mainly due to the skin deformation caused by carotid pulse. While these assumptions may not hold in certain real-life scenarios (e.g.,~talking, eating), more robust approaches can be considered to capture and compensate those motion sources. Similarly, our study showed promising results in a pool of 12 healthy individuals. While the performance was fairly consistent across participants, more studies will need to be performed including a larger and more varied pool of participants (e.g.,~unhealthy individuals, children, elderly people). 

\section{Conclusion}

We have described, implemented and evaluated a novel methodology for measuring HR and BR from NIR video recordings of the human neck, and demonstrated its accuracy under different lighting conditions. 
While this paper only addressed the recovery of HR and BR, other physiological parameters such as heart rate variability (HRV) could be captured using a similar methodology. As more and more devices with NIR cameras become available, this technology offers new opportunities to provide comfortable physiological measurements and long-term health monitoring. 

\section*{Acknowledgements}
This work is supported by the MIT Media Lab Consortium and the SDSC Global Foundation, Inc.

\printbibliography

\end{document}